\documentclass[runningheads]{llncs}
\usepackage{graphicx}
\usepackage{floatrow}
\newfloatcommand{capbtabbox}{table}[][\FBwidth]
\usepackage[ruled, lined, linesnumbered, commentsnumbered, longend]{algorithm2e} 
\usepackage{xcolor}

\usepackage{hyperref}
\usepackage{amsmath}
\usepackage{wrapfig}
\usepackage{multirow}
\usepackage{amssymb}

\DeclareMathOperator*{\argmax}{arg\,max}

\definecolor{newcolor}{rgb}{.8,.349,.1}

\newcommand\redsout{\bgroup\markoverwith{\textcolor{red}{\rule[0.5ex]{2pt}{0.4pt}}}\ULon}

\begin{document}
\title{Continual Active Learning for Efficient Adaptation of Machine Learning Models to Changing Image Acquisition}
%\title{Continual Active Learning for Efficient Scanner Adaptation of Machine Learning Models}
%
\titlerunning{Continual Active Learning for Adaptation to Changing Image Acquisition}
\author{Matthias Perkonigg \and
Johannes Hofmanninger \and
Georg Langs}
\authorrunning{M. Perkonigg et al.}
\institute{Department of Biomedical Imaging and Image-guided Therapy, Computational Imaging Research Lab (CIR), Medical University of Vienna, Austria\\
\email{\{matthias.perkonigg, georg.langs\}@meduniwien.ac.at}\\
\url{www.cir.meduniwien.ac.at}}
\maketitle              % typeset the header of the contribution
\vspace{-2mm}
\begin{abstract}
Imaging in clinical routine is subject to changing scanner protocols, hardware, or policies in a typically heterogeneous set of acquisition hardware. Accuracy and reliability of deep learning models suffer from those changes as data and targets become inconsistent with their initial static training set. Continual learning can adapt to a continuous data stream of a changing imaging environment. 
Here, we propose a method for continual active learning on a data stream of medical images. It recognizes shifts or additions of new imaging sources - \emph{domains} -, adapts training accordingly, and selects optimal examples for labelling. Model training has to cope with a limited labelling budget, resembling typical real world scenarios. We demonstrate our method on T1-weighted magnetic resonance images from three different scanners with the task of brain age estimation. Results demonstrate that the proposed method outperforms naive active learning while requiring less manual labelling.

\end{abstract}
 
\section{Introduction}
The frequently changing scanner hardware, imaging protocols, and heterogeneous composition of acquisition routines in the clinical environment hamper the longevity and utility of deep learning models. After initial training on a static data set they need to be continuously adapted to the changing characteristics of the data stream acquired in imaging departments. This is challenging, since training on a data stream without a continual learning strategy can lead to \textit{catastrophic forgetting} \cite{McCloskey1989}, when adapting a model to a new domain or task leads to a deterioration of performance on preceding domains or tasks. Additionally, for continual learning in a medical context manual labelling is required for new cases. However, labelling is expensive and time-consuming, requiring extensive medical knowledge. Here, active learning is a possible solution, where the aim is to identify samples from an unlabelled distribution that are most informative if labelled next. Keeping the number of cases requiring labelling as low as possible is a key challenge in active learning with medical images \cite{Budd2019AAnalysis}.

Most currently proposed continual active learning methods either do not take domain shifts in the training distributions into account or assume knowledge about the domain membership of data points. However, due to the variability in how meta data is encoded knowledge about the domain membership can not be assumed in clinical practice~\cite{Gonzalez2020WhatSegmentation}. Therefore, we need a technique to reliably detect domain shifts in a continuous data stream.
Combining continual and active learning with the detection of domain shifts can ensure that models perform well on a growing repertoire of image acquisition technology, while at the same time minimizing the resource requirements, and day to day effort necessary for continued model training. 

\begin{figure}[t]
    \centering
    \includegraphics[width=0.9\textwidth]{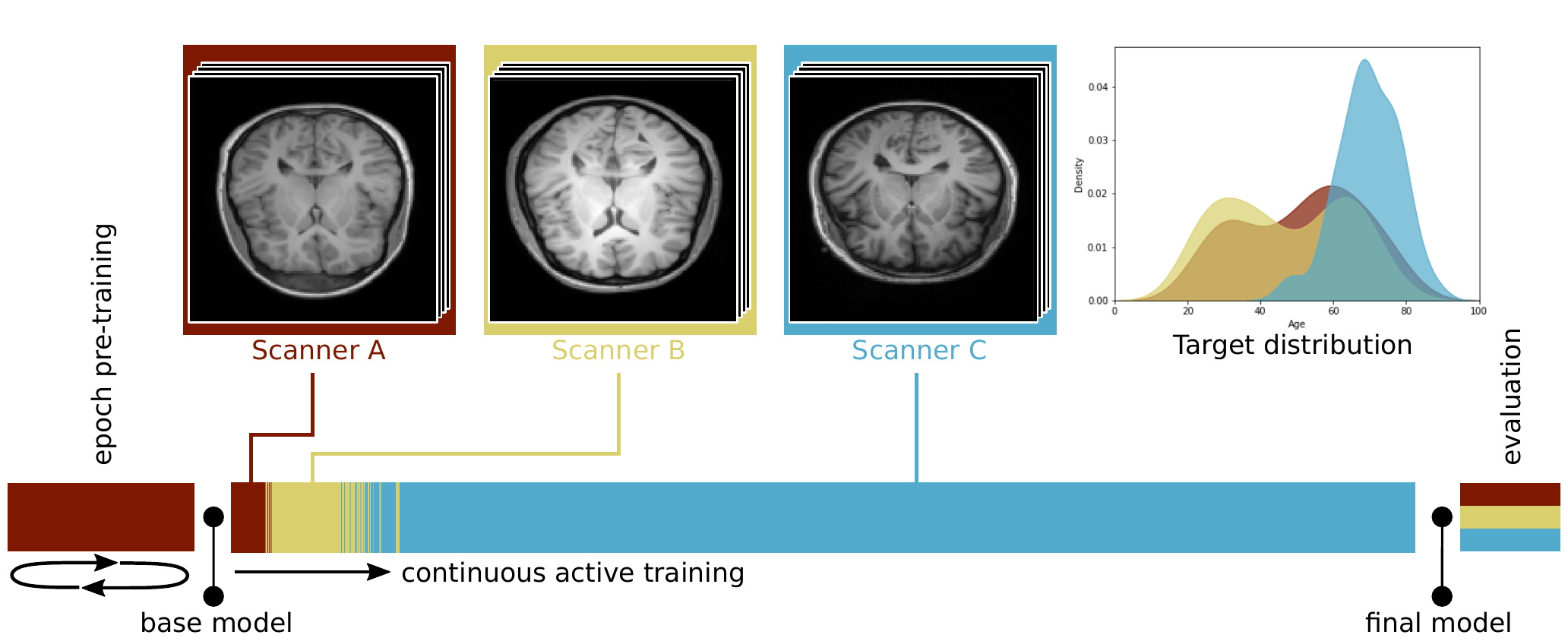}
    \caption{Experimental setup: A model is pre-trained on scanner A and then subsequently updated on a data stream gradually including data of scanner B and scanner C. The model has to recognize the domain shifts, and identify cases whose annotation will contribute best to the model accuracy. The final model is evaluated on test sets of all three scanners.}
    \label{fig:exp_setup}
\end{figure}

\paragraph{Contribution}  We propose an online active learning method in a setting where the domain membership (i.e. which scanner an image was acquired with) is unknown and in which we learn continuously from a data stream, by selecting an informative set of training examples to be annotated. Figure \ref{fig:exp_setup} illustrates the experimental setup reflecting this scenario. A model is pre-trained on data of one scanner. Subsequently, our method observes a continual stream and automatically detects if domain shifts occur, activating the learning algorithm to incorporate knowledge about the new domain in the form of labelled training examples. We evaluate the model on data from all observed domains, to assess if the model performs well on a specific task for all observed scanners. The proposed algorithm combines online active learning and novel domain recognition for continual learning. The proposed Continual Active Learning for Scanner Adaptation (CASA) approach uses a rehearsal method and an active labelling approach without any prior knowledge of the domain membership of each data sample. CASA learns on a stream of radiology images with a limited labelling budget, which is desirable to keep the required expert effort low.

\paragraph{Related work}
Our work is related to three research areas: First, continual learning with the goal to adapt a deep learning system to a continual stream of data while mitigating catastrophic forgetting. Second, active learning, which aims to wisely select samples to be labelled by an oracle to keep the number of required labels low. Third, domain adaptation, with the goal of adapting knowledge learned on one domain to a new target domain.
An overview of \textbf{continual learning} in medical imaging is given in \cite{Pianykh2020ContinuousApplications}. Here, we focus on the group of continual learning methods closest related to ours, namely \textit{rehearsal and pseudo-rehearsal methods}, which store a subset of training samples and periodically use them for training. Our memory interprets different domains as images having a different style, this is closely related to \cite{Hofmanninger2020DynamicSettings}. They use a dynamic memory which updates the subset for training without assuming domain knowledge based on a gram-matrix based distance. Furthermore, \cite{Ozdemir2018LearnImages} incrementally add different anatomy to a segmentation network by keeping a representative set of images from previous tasks and adding a new segmentation head for each new task.
\cite{Lenga2020} used the pseudo-rehearsal technique Learning-without-Forgetting (LwF) \cite{Li2018} on a domain adaptation problem on chest X-Ray.
A detailed review and discussion of \textbf{active learning} and human-in-the-loop systems in medical imaging can be found \cite{Budd2019AAnalysis}.
Also \cite{Pianykh2020ContinuousApplications} describes how continual learning approaches combined with human-in-the-loop algorithms can be useful in clinical practice. The setting in which our method operates is closely related to \textit{stream-based selective sampling} as described in \cite{Budd2019AAnalysis}, where a continual stream of unannotated data is assumed. There, the decision of whether or not labelling is required is made for every item on the stream independently. However, the authors claim that this family of methods have limited benefits due to the isolated decision for each sample. In this work, we alleviate this limitation by taken collected information about the data distribution observed into account.
In the area of \textbf{domain adaptation}, various approaches were proposed to adapt to continuously shifting domains. \cite{Wu2019ACE:Segmentation} proposed an approach for semantic segmentation in street scenes under different lightning conditions.  \cite{Lao2020ContinuousReplay,Bobu2018AdaptingDomains} proposed replay methods for continuous domain adaptation and showed performance in benchmark data sets (rotated MNIST \cite{Bobu2018AdaptingDomains} and Office-31 \cite{Lao2020ContinuousReplay} respectively).
In the area of medical imaging an approach for lung segmentation in X-rays with shifting domains was proposed in \cite{Venkataramani2019}.

\section{Method}
The proposed method CASA is a continual \textit{rehearsal method} with active sample selection.
We assume an unlabelled, continual image data stream $\mathcal{S}$ with unknown domain shifts. There exists an oracle that can be queried and returns a task label $y=\mathbf{o}(x)\ |\ x\in\mathcal{S}$. In a clinical setting this oracle would be a radiologist who labels an image. Since expert annotations are expensive, label generation by the oracle is limited by the label-budget $\beta$. The goal of CASA is to train a task network on such a data stream with the restriction of $\beta$, while mitigating catastrophic forgetting. We achieve that by detecting \textit{pseudo-domains}, defined as groups of images similar in terms of style. The identification of pseudo-domains helps to keep a diverse set of training examples. The proposed approach consists of two modules and two memories, which are connected by the CASA-Algorithm, described in the following.

\begin{figure}[t]
    \centering
    \includegraphics[width=0.8\textwidth]{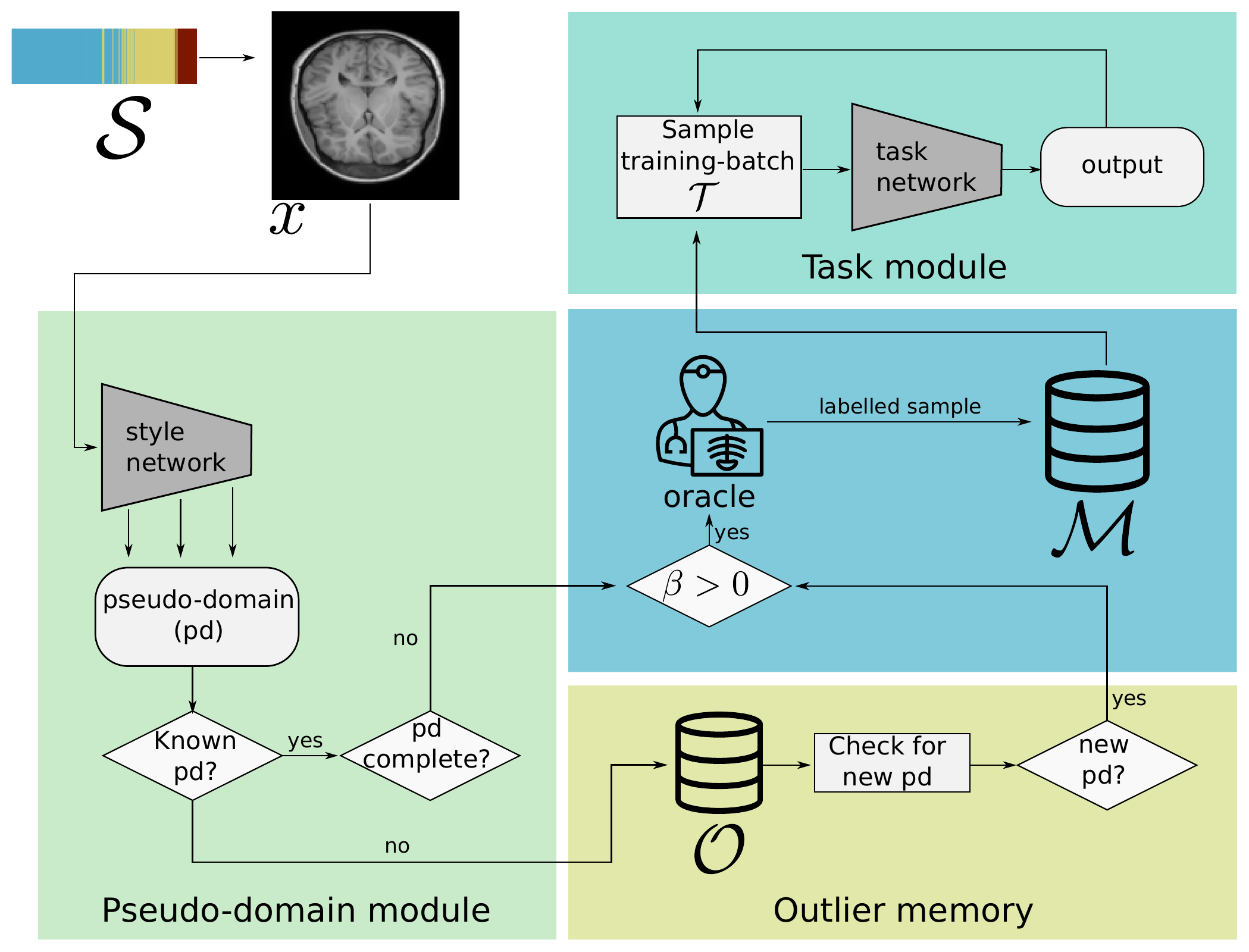}
    \caption{Overview of the \textit{CASA} algorithm. Each sample from the data stream is processed by the pseudo-domain module to decide whether its routed to the oracle or to the outlier memory. Whenever a new item is added to the outlier memory  it is evaluated if a new pseudo-domain (pd) should be created. The oracle labels a sample and stores it in the training memory, from which the task module trains a network. Binary decision alternatives resulting in discarding the sample are left out for clarity of the figure.}
    \label{fig:CASA_overview}
\end{figure}

\subsection{CASA Training Scheme}
Before starting continual training, the task module is pre-trained on a labelled data set $\mathcal{L} = \{\langle i_1,l_1\rangle, \dots, \langle i_L,l_L\rangle\}$, of image-label pairs $\langle i, l \rangle$ of a particular scanner. This pre-training is done in a conventional epoch based training procedure to start continual training on a model which performs well for a given scanner. 

After pre-training is finished, continual training follows the scheme depicted in Figure \ref{fig:CASA_overview}.
As a first step, an \textit{input-mini-batch} $\mathcal{B} = \{x_1, \dots, x_B\}$ is drawn from $\mathcal{S}$ and the \textit{pseudo-domain module} decides for each image $x\in \mathcal{B}$ to add it to one of the memories ($\mathcal{O}$ or $\mathcal{M}$) or discard the image. $\mathcal{M}$ is defined as a fixed size $M$ samples holding \textit{training memory} of labelled data $\mathcal{M} = \{\langle m_1,n_1,d_1\rangle \dots, \langle m_M,n_M, d_M\rangle\}$, where $m$ is the image, $n$ the corresponding task label and $d$ the \textit{pseudo-domain} the image was assigned to. Task labels $n$ are generated by calling the  \textit{oracle} $\mathbf{o}(x)$. Pseudo-domains $d$ are defined and assigned by the process described in Section \ref{sec:domain_assignment}. $\mathcal{M}$ is initialized with a random subset of $\mathcal{L}$ before starting continual training. The \textit{outlier memory} $\mathcal{O} = \{\langle o_1,c_1\rangle \dots, \langle o_n,c_n\rangle\}$ holds a set of unlabelled images. Here, $o$ is an image and $c$ is a count how long (measured in training steps) the image is a member of $\mathcal{O}$.

If there are pseudo-domains for which training is not completed, a training step is done by sampling a \textit{training-mini-batches} $\mathcal{T} = \{\langle t_1,{u}_1\rangle, \dots, \langle t_T,u_T\rangle\}$ of size $T$ from $\mathcal{M}$ and then training the task module for one step on $\mathcal{T}$. This process continues with the next $\mathcal{B}$ drawn from $\mathcal{S}$ until the end of $\mathcal{S}$.

\subsection{Pseudo-domain module}
The pseudo-domain module is responsible to evaluate the style of an image $x$ and assignment of $x$ to a pseudo-domain. Pseudo-domains are defined as groups of images that are similar in terms of style. Since our method does not assume direct domain (i.e. scanner and/or protocol) knowledge, we use the procedure described below to identify pseudo-domains as part of continual learning.
We define the set of pseudo-domains as $\mathcal{D} = \{\langle \mathbf{i_1}, \bar{p}_1 \rangle \dots \langle \mathbf{i_D}, \bar{p}_D\rangle\}$. Where $\mathbf{i_j}$ is a trained \textit{Isolation Forest (IF)} \cite{TonyLiu2008IsolationForest} used as one-class anomaly detection for the pseudo-domain $j\in \{1, \dots, D\}$. We use IFs, because of their simplicity and the good performance on small sample sizes. $\bar{p}_j$ is the running average of a classification or regression performance metric of the target task calculated on the last $P$ elements of pseudo-domain $j$ that have been labelled by the \textit{oracle}. The performance metric is measured before training on the sample. $\bar{p}_j$ is used to evaluate if the pseudo-domain completed training, that is $\bar{p}_{j}>k$ for classification tasks and $\bar{p}_{j}<k$ for regression tasks, where $k$ is a fixed performance threshold.
CASA then uses pseudo-domains to assess if training for a specific style is needed and to diversify the memory $\mathcal{M}$.

\subsubsection{Pseudo-domain assignment}
\label{sec:domain_assignment}
To assign an image to an existing pseudo-domain we use a \textit{style network}, which is pre-trained on a different dataset (not necessarily related to the task) and not updated during training. From this network we evaluate the style of an image based on the gram matrix $G^l \in \mathbb{R}^{N_l \times N_l}$, where $N_l$ is the number of feature maps in layer $l$. Following \cite{Gatys2016,Hofmanninger2020DynamicSettings} $G_{ij}^l(x)$ is defined as the inner product between the vectorized activations $\mathbf{f}_{il}(x)$ and $\mathbf{f}_{jl}(x)$ of two feature maps $i$ and $j$ in a layer $l$ given a sample image $x$:
\begin{equation}
\displaystyle G_{ij}^l(x) = \frac{1}{N_lM_l} \mathbf{f}_{il}(x)^\top\mathbf{f}_{jl}(x)
\label{eq:grammatrix}
\end{equation}
where $M_l$ denotes the number of elements in the vectorized feature map. 
Based on the gram matrix we define a \textit{gram embedding} $\mathbf{e}(x)$: For a set of convolutional layers $\mathcal{C}$ of the style network, we calculate the gram matrices ($G^l\ |\ l \in \mathcal{C}$) and apply a dimensionality reduction using Sparse Random Projection (SRP) \cite{Li2007VeryReduction}.
The isolation forests in $\mathcal{D}$ are trained on those embeddings $\mathbf{e}(x)$. We assign an image $x$ to the pseudo-domain maximizing the decision function of $\mathbf{i_d}$:
\begin{equation}
    \mathbf{p}(x) = 
\begin{cases}
    \underset{d}{\argmax}\ \mathbf{i_d}(\mathbf{e}(x)) & \text{if}\ \underset{d}{\text{max}}[\mathbf{i_d}(\mathbf{e}(x))]  > 0\\
    -1,              & \text{otherwise}
\end{cases}
|\ d \in \{1,\dots,D\}
\end{equation}

If $\mathbf{p}(x)=-1$ the image is added to the outlier memory $\mathcal{O}$ from which new pseudo-domains are detected (see Section \ref{sec:outlier_memory}). If the pseudo-domain $\mathbf{p}(x)$ is known and have completed training we discard the image, otherwise it is added to $\mathcal{M}$ according to the strategy described in Section \ref{sec:training_memory}.

\subsection{Task module}
The task module is responsible for learning the target task (e.g. age estimation in brain MRI), the main component of this module is the \textit{task network} ($\mathbf{t}(x)\mapsto y$), learning a mapping from input image $x$ to target label $y$. This module is pre-trained on a labelled data set $\mathcal{L}$. During continual training, the module is updated by drawing $n$ training-input-batches $\mathcal{T}$ from $\mathcal{M}$ and performing a training step on each of the batches. At the end of training the aim of CASA is that the task module performs well on images of all scanners available in $\mathcal{S}$ without suffering catastrophic forgetting.

\subsection{Training memory}
\label{sec:training_memory}
The $M$ sized training memory $\mathcal{M}$ is balanced between the pseudo-domains currently in $\mathcal{D}$. Each of the $D$ pseudo-domains can occupy up to $\frac{M}{D}$ elements in the memory. If a new pseudo-domain is added to $\mathcal{D}$ (see Section \ref{sec:outlier_memory}) a random subset of elements of all previous domain are flagged for deletion, so that only $\frac{M}{D}$ are kept protected in $\mathcal{M}$. 
If a new element $e=\langle m_k, n_k, d_k \rangle$ is to be inserted to $\mathcal{M}$ and $\frac{M}{D}$ is not reached an element currently flagged for deletion is replaced by $e$. Otherwise the element will replace the one in $\mathcal{M}$, which is of the same pseudo-domain and minimizes the distance between the gram embeddings. Formally we replace the element with index:
\begin{equation}
 \xi(i) = (\mathbf{e}(m_k)- \mathbf{e}(m_j))^2 |\ n_k = n_j,\ j \in \{1,\dots,M\}.
\end{equation}

\subsection{Outlier memory and pseudo-domain identification}
\label{sec:outlier_memory}
The outlier memory $\mathcal{O}$ holds candidate examples that do not fit an already identified domain, and might form a new domain. Examples are stored until they are assigned a new pseudo-domain or if a fixed number of training steps is reached. If no new pseudo-domain is discovered for an image it is considered a 'real' outlier and removed from the outlier memory. Within $\mathcal{O}$ we discover new pseudo-domains to add to $\mathcal{D}$. The discovery process is started when $\lvert \mathcal{O}\rvert=o$, where $o$ is a fixed threshold. A check if a dense region is present in the memory is done by calculating the pairwise euclidean distances of all elements in $\mathcal{O}$. If there is a group of images where the distances are below a threshold $t$ a new IF $\mathbf(i_n)$ is fitted to the gram embeddings of the dense region and the set of pseudo-domains $\mathcal{D}$ is updated by adding $\langle\mathbf{i_n}, \bar{p}_n\rangle$. For all elements belonging to the new pseudo-domain labels are queried from the oracle and they are transferred from $\mathcal{O}$ to $\mathcal{M}$.

\section{Experiments and Results}

We evaluated our method on the task of brain age estimation on T1-weighted magnetic resonance imaging (MRI) and compare CASA to different baseline techniques, described in Section \ref{sec:baselines}. 
First, we evaluate the task performance on all domains to show how the mean absolute error between predictions and ground truth changes on the validation set (Section \ref{sec:quantiative_results}). Furthermore, we evaluate the ability of continual learning to improve accuracy on existing domains by adding new domains \textit{backward transfer} (BWT), and the contribution of previous domains in the training data to improving the accuracy on new domains \textit{forward transfer} (FWT)\cite{Lopez-Paz2017}. BWT measure how learning a new domain influences the performance on previous tasks, FWT quantifies the influence on future tasks. Negative BWT values indicate catastrophic forgetting, thus avoiding negative BWT is especially important for continual learning.
In Section \ref{sec:memory_analysis}, we analyze the memory elements at the end of training to evaluate if the detected pseudo-domains match the real domains determined by the scanner types.

\subsection{Data set}
\begin{wraptable}{r}{7.5cm}
    \vspace{-2.7em}
    \setlength\intextsep{0pt}
    \centering
    \begin{tabular}{l|l|l|l|l}
        \centering

Scanner         &   1.5T IXI &  3.0T IXI &  3.0T OASIS &  Total \\
\hline \hline
Pre-train            &   201 &      0 &      0 &  201 \\
Continual       &   52 &   146 &   1504 &   1702\\
Validation       &    31 &    18 &    187 &   236\\
Test             &    31 &    18 &    187 &   236\\
        \hline
    \end{tabular}
    \caption{\textbf{Data:} Splitting of the data into a pre-train, continual, validation, and test set. The number of cases in each split are shown.}
    \label{tbl:data_brain}
    %\vspace{-2em}
\end{wraptable}
We use data pooled from two different data sets containing three different scanners. We use the IXI data set\footnote{https://brain-development.org/ixi-dataset/} and data from OASIS-3 \cite{LaMontagne2019OASIS-3:Disease}. From IXI we use data from a Philips Gyroscan Intera 1.5T and a Philips Intera 3.0T scanner, from OASIS-3 we use data from a Siemens TrioTim 3.0T scanner. Data was split into base pre-train, continual training, validation and test set (see Table \ref{tbl:data_brain}). Images are resized to 64x128x128 and normalized to a range between 0 and 1.

\subsection{Methods compared in the evaluation}
\label{sec:baselines}
We compared four methods in our evaluation:
\begin{enumerate}
    \item \textit{Joint model}: a model trained in a standard, epoch-based approach on samples from all scanners in the experiment jointly.
    \item \textit{Scanner models}: a separate model for each scanner in the experiment trained with standard epoch-based training. The evaluation for a scanner is done for each scanner setting separately.
    \item \textit{Naive AL}: a naive continuously trained, active learning approach of labelling every $n$-th label from the data stream, where $n$ depends on the labelling budget $\beta$.
    \item \textit{CASA (proposed method)}: The method described in this work. The settings for experimental parameters for CASA are described in Section \ref{sec:brainage_exp}.
\end{enumerate}

The joint and scanner specific models (1, 2) require the whole training data set to be labelled and available at once, thus they are an upper limit to which our method is compared to. Naive AL and CASA use an oracle to label specific samples only. The comparison with naive AL evaluates the gains of our method of choosing samples to label by detecting domain shifts. Note, that as the goal of our experiments is to show the impact of CASA in comparison to baseline active continual learning strategies and not to develop the best network for brain age estimation we do not compare to state-of-the-art brain age estimation methods.

\subsection{Experimental setup}
\label{sec:brainage_exp}
The task model is a simple feed-forward network as described in \cite{Dinsdale2020UnlearningSegmentation} and provided on github\footnote{https://github.com/nkdinsdale/Unlearning\_for\_MRI\_harmonisation}. The style network used in the pseudo-domain module is a 3D-ModelGenesis model pre-trained on computed tomography images on the lung \cite{Zhou2020ModelsGenesis}.
We run experiments with different parameter settings evaluating the influence of the memory size $M$, the task performance threshold $k$, and the labelling budget $\beta$, expressed as a fraction of the continuous training set. We test the influence of $\beta$ with different settings $\beta=\frac{1}{20}$ (n=85), $\frac{1}{10}$ (n=170), $\frac{1}{8}$ (n=212) and $\frac{1}{5}$ (n=340). For the performance threshold, we tested $k=5.0$ and $k=7.0$, to demonstrate the influence of $k$ on the labelling need of CASA. Values for $k$ are set after observing the performance of the baseline models.
The main performance measure for brain age estimation we use is the mean absolute error (MAE) between predicted and true age.

\subsection{Model Accuracy Across Domains}
\label{sec:quantiative_results}
\begin{figure}[t]
    \centering
    \includegraphics[width=0.9\textwidth]{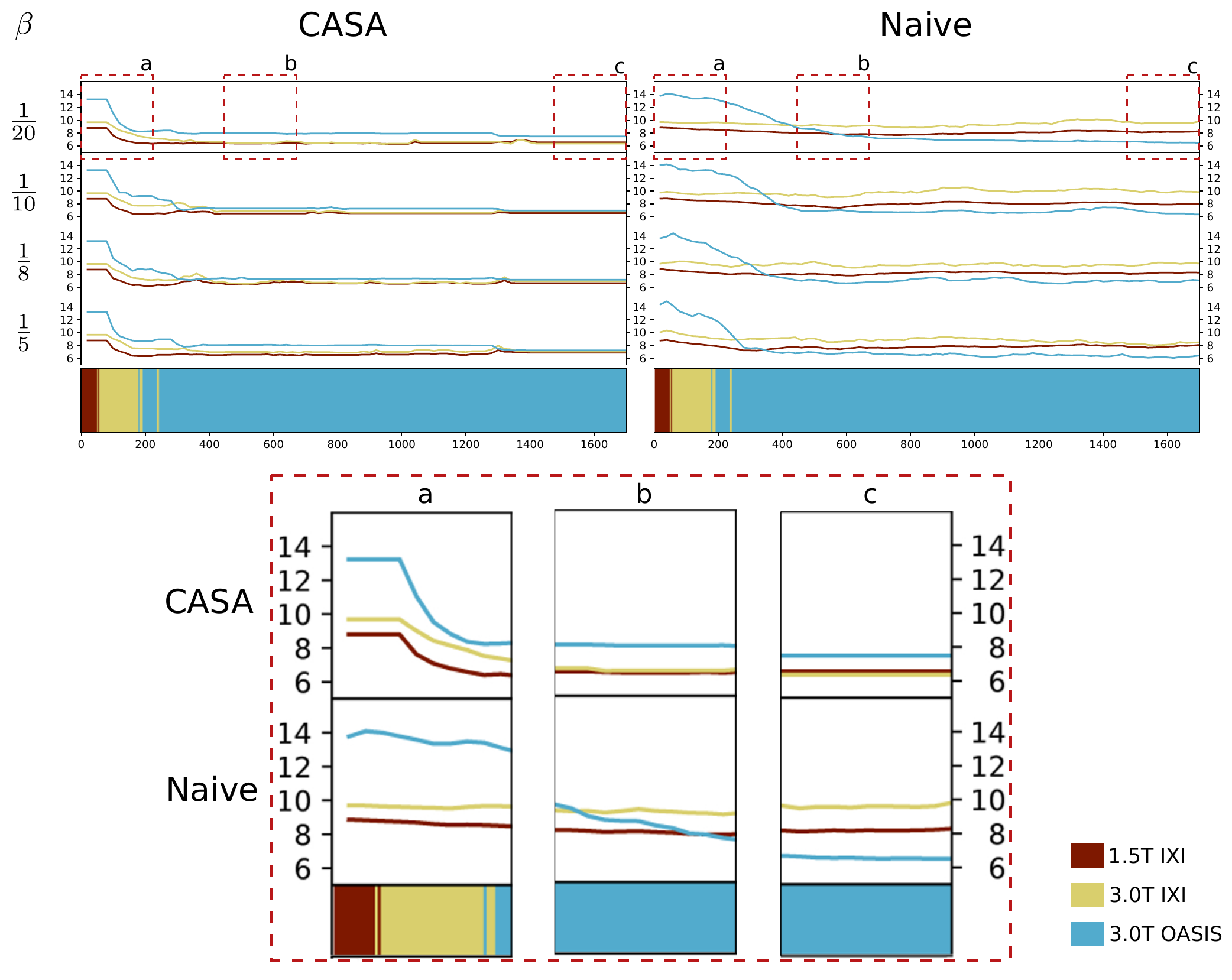}
    \caption{Evaluation for $M=128$ and $k=5.0$ with different $\beta$. Y-axis shows the mean absolute error (MAE, lower is better) of the model on the validation set. Zoomed in sections of training steps of particular interest for $\beta=\frac{1}{20}$: (a): CASA detects the domain shift to 3.0T IXI and trains on the new domain, this also leads to a big forward transfer for 3.0T OASIS. Naive AL only takes every 20-th image for training, thus failing to adapt to 3.0T IXI. (b): CASA is relatively stable, while naive AL incorporates more knowledge about 3.0T OASIS images and start to overfit on those, while showing slight forgetting for 1.5T IXI and 3.0T IXI (c): At the end of the continuous stream CASA show an equal performance for all three scanners, while naive AL leads to good performance on the last domain, but significantly poorer results on previous domains.}
    \label{fig:brain_val_results}
\end{figure}

Figure \ref{fig:brain_val_results} shows how the mean absolute error on the validation set, of CASA and the naive AL approach changes during training. Adaption to new domains is much faster when using our approach, even if the labeling budget is low (e.g. $\beta=\frac{1}{20}$). Furthermore, as seen from the curves training is more stable. Lower values are reached for CASA in comparison to naive AL at the end of training across all scanners for the validation set.

In Table \ref{tab:results_brain} different parameter settings are compared. The performance gap between CASA and naive AL trained with $M=128$ is especially large for images of 3.0T IXI. There, our approach successfully identified the new domain as a pseudo-domain, and trained the model accordingly. Naive AL takes every $n$-th element according to $\beta$, thus samples from scanner 3.0T IXI are seen less often and the model cannot adapt to this domain. The advantage of CASA increases as the labelling budget is reduced. Evaluating two different memory sizes $M=64$ and $M=128$ shows, that CASA could not gain performance over a naive AL approach when $M$ is small.
Comparing CASA with $k=7.0$ to $k=5.0$ demonstrates, that a more challenging choice of $k$ leads to better overall performance, furthermore CASA takes more advantage of a larger $\beta$ when $k=5.0$. 

\begin{table}[t]
\centering
\begin{tabular}{c|c|l|c|c|l|l|l||l|l}
$\beta$ &  Labelled &    Meth.   & $M$ & $k$ & 1.5T IXI & 3.0T IXI & 3.0T OASIS &BWT&FWT\\
\hline
\multirow{3}{*}{$\frac{1}{20}$}     & [74-85]   & CASA      & 128 & 5.0 & $6.29\pm0.26$ & $6.22\pm0.54$ & $6.51\pm0.81$ & $0.98\pm0.15$ & $6.06\pm0.78$\\
                                    & [70-85]   & CASA      & 128 & 7.0 & $6.11\pm0.22$ & $7.27\pm0.14$ & $6.24\pm0.20$ & $0.87\pm0.27$ & $5.57\pm0.65$\\
                                    & 85        & NAL  & 128 & -   & $6.89\pm0.28$ & $9.88\pm0.49$ & $5.98\pm0.11$ & $0.76\pm0.28$ & $3.69\pm0.72$\\\hline
\multirow{3}{*}{$\frac{1}{10}$}     & [90-170]  & CASA      & 128 & 5.0 & $6.03\pm0.40$ & $6.57\pm0.58$ & $5.75\pm0.47$ & $1.07\pm0.42$ & $5.57\pm0.86$\\
                                    & [74-75]   & CASA      & 128 & 7.0 & $5.89\pm0.07$ & $6.93\pm0.39$ & $6.45\pm0.12$ & $0.79\pm0.10$ & $5.84\pm0.92$\\
                                    & 170       & NAL  & 128 & -   & $6.71\pm0.28$ & $9.08\pm0.23$ & $5.72\pm0.15$ & $1.06\pm0.23$ & $3.93\pm0.81$\\\hline
\multirow{3}{*}{$\frac{1}{8}$}      & [91-212]  & CASA      & 128 & 5.0 & $6.65\pm0.28$ & $6.38\pm0.89$ & $5.86\pm0.45$ & $0.78\pm0.46$ & $6.05\pm0.96$\\
                                    & [74-75]   & CASA      & 128 & 7.0 & $6.14\pm0.12$ & $7.34\pm0.13$ & $6.88\pm0.63$ & $0.52\pm0.23$ & $5.55\pm0.77$\\
                                    & 212       & NAL  & 128 & -   & $7.02\pm0.17$ & $9.29\pm0.23$ & $6.64\pm0.27$ & $0.81\pm0.08$ & $4.07\pm0.85$\\\hline
\multirow{3}{*}{$\frac{1}{5}$}      & [91-112]  & CASA      & 128 & 5.0 & $6.20\pm0.28$ & $6.68\pm0.33$ & $5.94\pm0.20$ & $0.74\pm0.03$ & $5.94\pm0.79$\\
                                    & [69-110]  & CASA      & 128 & 7.0 & $5.99\pm0.16$ & $7.09\pm0.45$ & $6.16\pm0.29$ & $1.07\pm0.06$ & $5.64\pm0.92$\\
                                    & 340       & NAL  & 128 & -   & $6.34\pm0.40$ & $8.06\pm0.28$ & $5.57\pm0.13$ & $0.90\pm0.32$ & $4.27\pm1.09$\\\hline \hline
\multirow{2}{*}{$\frac{1}{20}$}     & [62-66]   & CASA      & 64 & 7.0 & $6.70\pm0.52$ & $9.51\pm0.67$ & $5.95\pm0.05$ & $1.30\pm0.51$ & $3.39\pm0.69$\\
                                    & 85        & NAL     & 64 & - & $6.53\pm0.23$ & $9.49\pm0.26$ & $6.20\pm0.33$ & $1.30\pm0.51$ & $3.74\pm0.70$\\\hline
\multirow{2}{*}{$\frac{1}{10}$}     & [63-69]   & CASA      & 64 & 7.0 & $6.61\pm0.29$ & $9.07\pm0.33$ & $5.99\pm0.19$ & $1.54\pm0.25$ & $3.47\pm0.71$\\
                                    & 170       & NAL  & 64 & - & $6.39\pm0.32$ & $8.57\pm0.44$ & $5.61\pm0.15$ & $1.41\pm0.19$ & $4.00\pm1.01$\\\hline
\multirow{2}{*}{$\frac{1}{8}$}      & [63-64]   & CASA      & 64 & 7.0 & $6.21\pm0.17$ & $9.00\pm0.06$ & $6.01\pm0.23$ & $1.83\pm0.05$ & $3.39\pm0.72$\\
                                    & 212       & NAL  & 64 & - & $6.64\pm0.22$ & $9.18\pm0.10$ & $6.49\pm0.22$ & $1.27\pm0.16$ & $3.80\pm0.76$\\\hline
\multirow{2}{*}{$\frac{1}{5}$}      & [62-67]   & CASA      & 64 & 7.0 & $6.44\pm0.54$ & $9.57\pm0.44$ & $5.77\pm0.30$ & $1.38\pm0.11$ & $3.42\pm0.83$\\
                                    & 340       & NAL  & 64 & - & $6.09\pm0.16$ & $8.32\pm0.07$ & $5.10\pm0.13$ & $0.98\pm0.14$ & $4.21\pm0.83$\\\hline\hline
& & ScM & & & $6.61\pm0.71$ & $8.22\pm0.95$ & $6.62\pm0.97$\\
& & JM & & & $9.88\pm0.30$ & $7.14\pm1.35$ & $5.85\pm0.33$ \\ \hline
\end{tabular}
\caption{Results for age estimation on a test set reported as mean absolute error (MAE, lower is better, $\pm$ indicates the interval of results with 3 independent runs with different seeds). The table compares CASA, naive active learning (NAL), individual scanner models (ScM), and a joint model trained from all data (JM). The column \textit{Labelled} denotes the amount of labelling by the oracle needed during training.}
\label{tab:results_brain}
\end{table}

The BWT and FWT comparison in Table \ref{tab:results_brain} shows positive values for all approaches. Since the task remains the same and only the imaging domains change, this is an expected behaviour. Backward transfer is similar between CASA and naive AL over all settings. For approaches with $M=128$ we see a clear FWT gap between CASA and naive AL. This shows that CASA is able choose meaningful samples that are also helpful for subsequent tasks. 

\begin{figure}
    \includegraphics[width=0.8\textwidth]{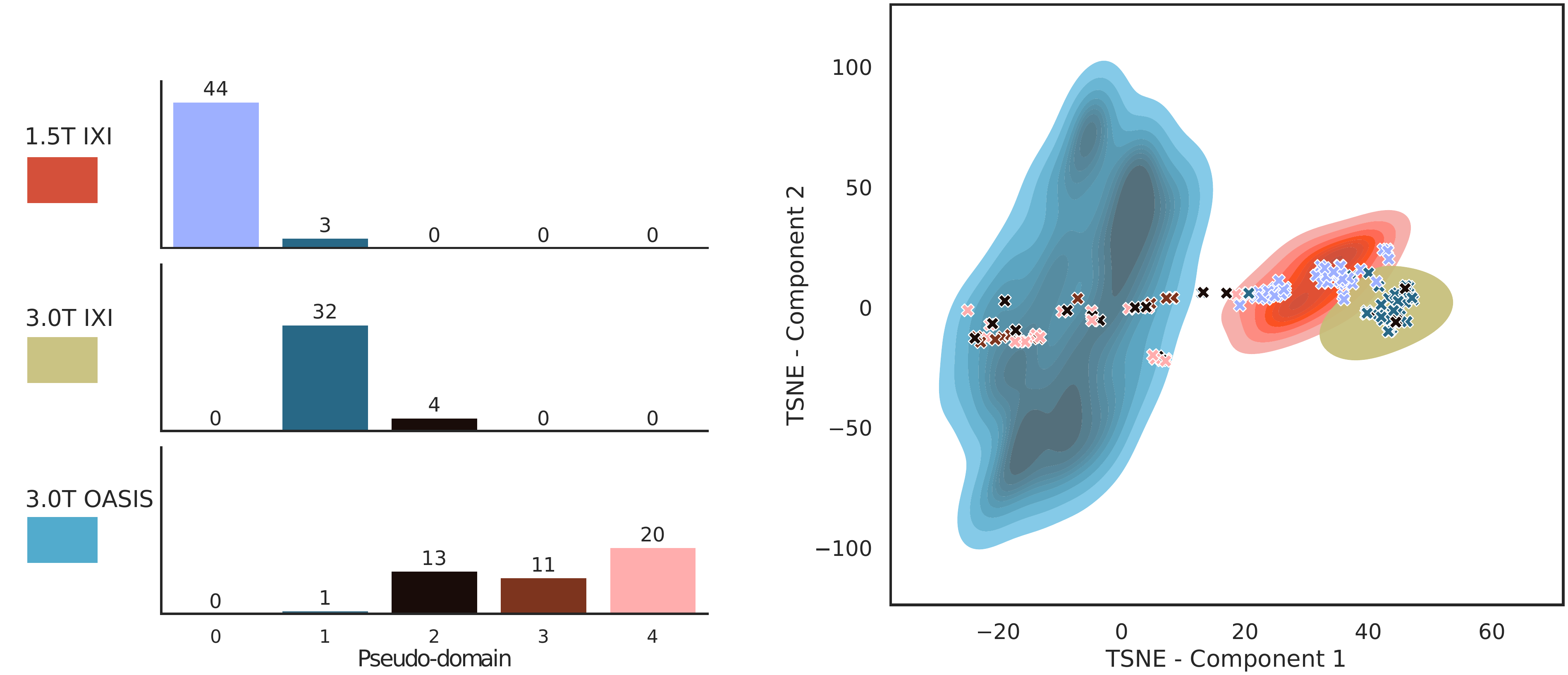}
    \caption{\textbf{Detected pseudo-domains capture scanner differences}: \textit{Left:} Distribution of pseudo-domains in $\mathcal{M}$ after training over 3 scanners. \textit{Right:} TSNE embedding of the gram embeddings for the whole training set, with marked position of elements in $\mathcal{M}$.}
    \label{fig:memory_analysis}
\end{figure}

\subsection{Evaluation of the Memory and Pseudo-Domains}
\label{sec:memory_analysis}
Here, we analyze the memory for the parameter settings $\beta=\frac{1}{10}$, $k=5.0$ and $M=128$. Other parameter settings show similar trends.
The final training memory $\mathcal{M}$ with CASA consists of 47 1.5T IXI, 36 3.0T IXI and 45 3.0T OASIS images. In comparison for naive AL 96 1.5T IXI, 7 3.0T IXI and 25 3.0T OASIS images are stored in the memory.
This demonstrates that with CASA the memory is more diverse and captures all three scanners equally, while for naive AL images from 3.0T IXI are heavily underrepresented. Detecting and balancing between the pseudo-domains as done in CASA is beneficial to represent the diversity of the training data.

Figure \ref{fig:memory_analysis} illustrates the capability of pseudo-domain detection to match real domains. We detect 5 pseudo-domains with the parameter setting mentioned above. Each pseudo-domain is present mainly in a single scanner, with the third scanner being represented by three pseudo-domains (2, 3 and 4). This might be related to appearance variability within a scanner. We plot a t-distributed Stochastic Neighbor Embedding (TSNE) \cite{maaten2008visualizing} for the gram embeddings of all samples in the training set. The shaded areas represent the density of samples from the individual scanners. The markers show where the the elements $m\in\mathcal{M}$ are located in the embedding of the whole data set and their assignments to pseudo-domains. This embedding demonstrates that samples in the final $\mathcal{M}$ are distributed over the embedding space, with areas with high sampling density for the under represented scanner 3.0T IXI and less dense areas for 3.0T OASIS with more samples.
\section{Conclusion}
We propose a continual active learning method for the adaptation of deep learning models to changing image acquisition technology. It detects emerging domains corresponding to new imaging characteristics, and optimally assigns training examples for labeling in a continuous stream of imaging data. Continual learning is necessary to enable models that cover a changing set of scanners. Experiments show that the proposed approach improves model accuracy over all domains, avoids catastrophic forgetting, and exploits a limited budget of annotations well. 
\subsubsection*{Acknowledgments}
\vspace{-1em}
This work was supported by Novartis Pharmaceuticals Corporation and received funding by the European Union's Horizon 2020 research and innovation programme under the Marie Marie Sk\l odowska-Curie grant agreement No 765148.

\bibliographystyle{splncs04}
\bibliography{activecontinual}
\end{document}